\documentclass[10pt,twocolumn,letterpaper]{article}

\usepackage{iccv}
\usepackage{times}
\usepackage{epsfig}
\usepackage{graphicx}
\usepackage{amsmath}
\usepackage{amssymb}

\usepackage[pagebackref=true,breaklinks=true,letterpaper=true,colorlinks,bookmarks=false]{hyperref}

 \iccvfinalcopy 

\def\iccvPaperID{****} 

\ificcvfinal\fi
\begin{document}

\title{A State Of the Art Report on Research in Multiple RGB-D sensor Setups}

\author{Kai Berger\\
Oxford e-Research Centre\\
University Of Oxford\\
{\tt\small kai.berger@oerc.ox.ac.uk}
}

\maketitle

\begin{abstract}
That the Microsoft Kinect, an RGB-D sensor, transformed the gaming and end consumer sector has been anticipated by the developers. That it also impacted in rigorous computer vision research has probably been a surprise to the whole community. Shortly before the commercial deployment of its successor, Kinect One, the research literature fills with resumees and state-of-the art papers to summarize the development over the past 3 years. This particular report describes significant research projects which have built on sensoring setups that include two or more RGB-D sensors in one scene.
\end{abstract}

\section{Introduction}
\begin{figure*}
\centering
\includegraphics[height=.24\linewidth]{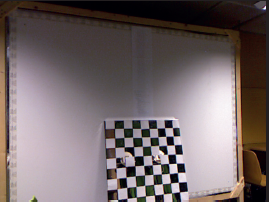}
\includegraphics[height=.24\linewidth]{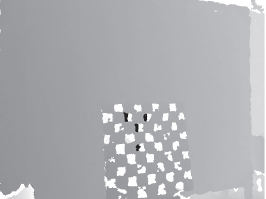}
\includegraphics[height=.24\linewidth]{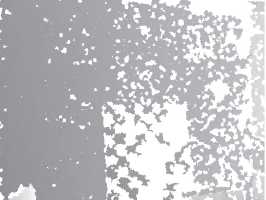}
\caption{\label{fig:multiple}A simple scene (left) captured with the depth camera of one (middle) and multiple concurrently projecting Kinects (right). The interference of more than one Kinect pattern results in degradations
in the captured depth image (white pixels denote invalid depth values). This state of the art report lists significant papers that implemented setups albeit interference issues or to specifically address and overcome these issues. Reproduced from Schroeder \etal \cite{schroder2011multiple}.}
\end{figure*}

With the release of the Microsoft Kinect in November 2010, Microsoft predicted a significant change in the use of gaming devices in the end consumer market. After a preview at the E3 game convention in the Windows Media Centre Environment, the selling in North America started at November 4, 2010 and up to today more than 24 million units have been sold. With the release of an open-source SDK named \textit{libfreenect} by H\`{e}ctor Mart\`{i}n that enables streaming both the depth and the RGB or the raw infrared images via USB the attention of young researchers to use the Microsoft Kinect sensor for their imaging and reconstruction applications has gained.
It was possible to stream 1200x960 RGB and IR images at a framerate of 30Hz alongside computed depth estimates of the scene at a lower resolution. The IR image featured the projected infrared pattern generated with an 830nm laser diode, which is distinctive and the same for each device.
Shortly thereafter the proceedings and journals in the community included papers describing a broad range of setups addressing well-known problems in computer vision in which the Microsoft RGB-D sensor was employed. The projects ranged from SLAM over 3d reconstruction over realtime face and hand tracking to motion capturing and gait analysis.
Counter-intuitively researchers became soon interested in addressing the question if it is possible to employ several Microsoft Kinects, i.e. RGB-D sensors, in one setup - and if so, how to mitigate interference errors in order to enhance the signal. This idea is mainly counter-intuitive due to the fact, the each device projects the same pattern at the same wavelength into the scene. Thus, one would expect that the confusion in processing the raw IR-data rises quickly with the amount of sensors installed in a scene, Fig.~\ref{fig:multiple}. In the following sections we give an overview over several research projects published in the proceedings and journals of the computer vision community that successfully overcome this preconception and highlight their challenges as well as the benefit of each multiple RGB-D sensor setup. A tabular overview about addressed papers is found in Table~\ref{overviewtable}.
\subsection{Method Of Comparison}
As this paper is a state-of-the art report it explicitly provides no new research contribution. Instead it shall be read as an overview and introduction to the work that has been conducted in the subfield of multiple Kinect research. We want to provide a comparative table, Table~\ref{overviewtable}, to have a short index of examined papers and the properties. The table is sorted alphabetically for each research field, i.e. \emph{Multiple RGB-D sensor Setups for Motion Estimation}, Sect.~\ref{motion}, \emph{Multiple RGB-D sensor Setups for Reconstruction}, Sect.~\ref{recon}, \emph{Multiple RGB-D sensor Setups for Recognition and Tracking}, Sect.~\ref{track}, and \emph{Interference in Multiple RGB-D sensor Setups}, Sect.~\ref{interference}. We compared the amount of Kinects installed in each capturing environment (third column), and stated where the sources were available the measured accuracy of the capturings. As the statements were not unified, we have to provide them in different units to adhere to the source text. A slightly more detailed description is given at the table caption. Finally we state if the capturing setup was externally calibrated to a common worldspace, usually performed with a checkerboard or moving a marker around the scene. 

\section{Multiple RGB-D sensor Setups for Motion Estimation}
\label{motion}
Santhanam \etal \cite{santhanam2011th} describe a system to track neck and head movements with four calibrated Kinects.
Three Kinects are tracking the patient’s
anatomy contour in depth and RGB streams while the fourth camera detects the face of the patient.
The detected face region is used to guide the contour detection in the other three views. The detected contours are then
finally merged to to a 3d estimate of the pose of the anatomy. The authors claim a precision of $3 mm$ at the expected $30 Hz$.
Wilson \etal \cite{wilson2010combining} use three PrimeSense depth cameras which stream at $320\times240px$ resolution and $30Hz$ for human interaction with an augmented reality table. They compare input depth image streams against background depth images for each depth camera captured when the room is empty to segment out the human user. While the authors do not specify the accuracy, e.g. between the projected area and the captured area comprised by a hand, they claim to robustly track all user actions in 10 cm volume above the table. The depth cameras were juxtaposed next to each other and slanted such that each camera captures a different angle in the room. However their viewing cones may have overlapped. Fuhrmann \etal \cite{fuhrmannmulti} have employed a stage setup with three Kinects for musical performances. They calibrated the cameras, which were observing the same $3\times3\times3 m^3$ interaction volume from different angles, for each stage performance. The tracking via \emph{OpenNI} suffered only from latency between interframe capturing times. The sensors were employed such that they did not interfere destructively.
Berger \etal \cite{berger2011markerless} employ four Kinect sensors in a small $3\times3\times3 m^3$ room to mitigate shortcomings in the motion capturing capabilities of a single Kinect, Fig.~\ref{fig:scenarios} (left). To overcome depth map degradation through interfering patterns they introduced external hardware shutters. The idea was further evaluated by Zhang \etal \cite{zhang2012real} who basically performed the same capturing only with two Kinect cameras. Interference issues were circumvented by placing them opposite each other and assuming that the human actor acts as a separation surface between both projection cones. The authors claim a tracking accuracy of $20cm$. Their processing algorithm limits the original capturing framerate of $30Hz$ to $15Hz$. Asteriadis \etal \cite{asteriadis2013estimating} included a treadmill to simulate partially occluded motion for three calibrated Kinect sensors placed evenly in a quarter arc around the treadmill. Using a Fuzzy Inference system they were able to robustly map the human motion. Although they do neither state reprojection errors nor deviations from a reconstructed mesh they provide figures that the human motion could be fitted by a skeleton in up to $95\%$ of the recorded frames. An approach to analyse facial motion with two Kinects is presented by Hossny \etal~\cite{hossny2012low}. They also provide a smart algorithm to automatically calibrate one Kinect to another based one rotation to zero angular positions. The processing of the depth maps to the the face is done with geometric features that outperform conventional Haar features. They propose to overcome interference difficulties with mutually rotated polarization filters but do not state figures about the reprojection error. Very recently, Ye \etal \cite{ye2013free} provided a solution for capturing human motion with multiple moving Kinects. The details about the number of Kinects used in the setup is currently not known to the authors of this report. Also, no knowledge about figures of Reconstruction errors exists currently.

\section{Multiple RGB-D sensor Setups for Reconstruction}
\label{recon}
\begin{figure*}
\centering
\includegraphics[width=.9\linewidth]{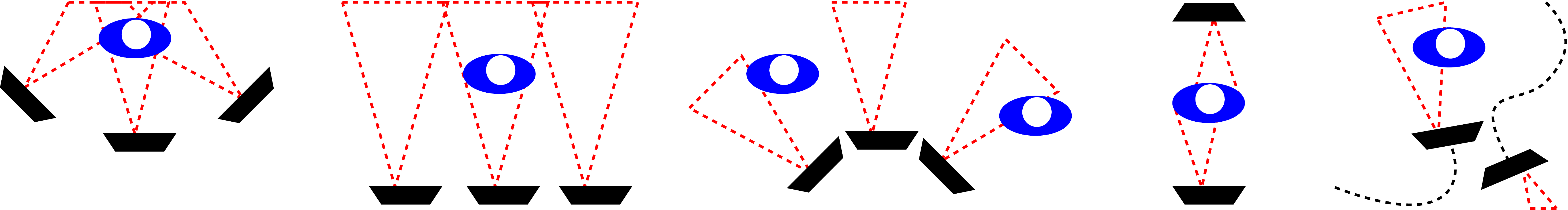}
\caption{\label{fig:scenarios}Five typical capturing setups featuring multiple Kinects. Multiple Kinects are evenly placed in a virtual circle around the scene centre (first), e.g. \cite{berger2011markerless,berger2011capturing,olesen2012real,wang20123d,nakazawa2012dynamic,satyavolu2012analysis,kainz2012omnikinect}, multiple Kinects are in line to capture a volume with a large side length (second), e.g. \cite{macknojia2013calibration,lothree,caon2011context,maimone2012reducing,saputra2012indoor}, multiple kinects juxtaposed and facing away from each other (third), e.g. \cite{wilson2010combining}, and two Kinects face each other, but are occluded by the scene content (fourth), e.g. \cite{nakamuramultiple}. Very recently work has been conducted with multiple uncalibrated moving Kinects (fifth), e.g. \cite{pancham2012mapping,ye2013free}}
\end{figure*}

Alexiadis \etal \cite{alexiadis2012reconstruction} use four Kinect devices to reconstruct a single,
full 3D textured mesh of a human body from their depth data in realtime. The authors claim that the re-projection error is less than 0.8 pixels. In a merging step redundant triangles are clipped. Object boundary noise is removed with a distance-to-background map. Rafibakhsh \etal \cite{rafibakhsh2012analysis} analyse construction site scenarios with two Kinects and exhaustively search for optimal placement an angles, concluding that the two sensors should not directly face each other. In their calibrated sensor setup they found a scene accuracy of $3.49 cm$.
Sumar \etal \cite{sumarfeasability} test the sensor interference for two uncalibrated Kinect sensors in an indoor environment. They found, that in a marker tracking task, where the markers are less than 3 meters from the Kinect the error follows a Gaussian distribution and does not deviate more than 5 pixels from the true centre of the marker. In ongoing work Pancham \etal \cite{pancham2012mapping} mount Kinects atop mobile robots which move in an overcast outdoor environment in order to segment out moving objects from static scenery. In that context the Kinect is used for differentiation between moving and stationary objects, and for map construction of the environment. They however do not state the accuracy of the reconstructed scene in relation to the amount of Kinects employed. In a very interesting approach to enable HDR scene capturing Lo \etal \cite{lothree} juxtapose two Kinects atop each other and equip one with a polarized neutral density filter resulting in accurate depth values for regions that would have been overexposed in an unaltered Kinect capturing (The exposure difference between both IR images is roughly 1 EV apart). They recognise the fact that interference might occur but did not quantitatively evaluate that for their setup. However, the reconstructed scenes bear more complete meshes under headlight than with a single LDR capturing.
Berger \etal \cite{berger2011capturing} show in their paper the feasibility to use three Kinects concurrently in a convergent setup for capturing non-opaque surfaces like the interface between flowing propane gas in air. It is noteworthy that, although the projectors are masked such that they project on mutually disjoint surface areas, the projection patterns do not interfere destructively with each other while passing through the gas volume. Their approach has been altered such that an evaluation based only on the high resolution IR stream is possible as well \cite{berger2013a}. Olesen \etal \cite{olesen2012real} show a system that involves up to three calibrated Kinects for texlet reconstruction. They evaluate different angular settings for the multiple sensors but interestingly conclude that the orientation does not significantly improve the capturing quality. In industrial applications Macknojia \etal \cite{macknojia2013calibration} juxtapose three Kinects on a straight line next to each other while a fourth and a fifth Kinect are placed to the left and right respectively in a convergent manner to provide a calibrated capturing volume with a side length of $7m$ in total, Fig.~\ref{fig:scenarios} (middle). Small projecting volumes overlap while objects like cars are captured. The authors state a depth error of about
$2.5 cm$ at $3m$ distance. Wang \etal \cite{wang20123d} present work where two calibrated Kinects' depth maps are fused to reconstruct arbitrary scene content. The cameras are spaced $30cm$ apart and the viewing axes converge towards the scene centre. Inaccuracies due to interference are handled in software by applying a his work Naveed \cite{ahmedsystem} provides a scene reconstruction mainly of human bodies captured from 6 calibrated Kinects. He deliberately excludes interference analysis from the discussion but mentioned temporal drift if software synchronization is omitted. Interference issues are also neglected by Nakazawa \etal \cite{nakazawa2012dynamic} who placed four calibrated Kinects at the four corners of a capturing room, but rotated them by $90^\circ$ such that they would capture a greater vertical range and a smaller horizontal range each. They concentrate on aligning depth data captured asynchronously by applying a temporal calibration by providing depth data at certain time instants. In their work Nakara \etal \cite{nakamuramultiple} place two Kinects in different angles between $10^\circ$ and $180^\circ$ from each other around the scene. The Kinects are not calibrated to a common world space but placed at a fixed distance to the scene centre. In an evaluation of the mean reprojection error for the varying angles they find that a spacing of $180^\circ$ between each Kinect results in the smallest error while a a spacing of $120^\circ$ results in the largest error, Fig.~\ref{fig:scenarios} (right). The Kinects do not project into each others sensor due to the scene content.
\section{Multiple RGB-D sensor Setups for Recognition and Tracking}
\label{track}
Satta \etal \cite{sattareal} present research to recognize and track people in an indoor environment surveyed by two Kinects relying on a combination of RGB texture and depth information.
It has to be noted, though, that the Kinects were installed facing away from each other. Hence, they did not directly project into each other's viewing frustra. Interference is not discussed further. Satyavolu \etal \cite{satyavolu2012analysis} describe an experimental setup that consists of 5 Kinects. One camera was used for tracking IR markers attached to a box, 4 others (evenly distributed around the scene centre) simulated interference/noise. The authors report that the Kinect deviated by $3cm$ on an average from the actual position. Caon \etal \cite{caon2011context} present an approach for tracking gestures based on three calibrated Kinects placed in a $45^\circ$ angle. They varied different configurations between the three Kameras and although they did not state figures about the depth or tracking accuracy they do list the amount of invalid depth pixels for each configuration.
Susanto \etal \cite{susanto20123d} present an approach to detect objects from their shape and depth profile generated when captured from several calibrated Kinects and state that there is no degrading interference noticeable due to the fact the the Kinects are placed at wide angles from each other. Although the paper focus on the success rate of the recognition they briefly state that the setup might show  depth discrepancies of up to $13 cm$.
The tracking of humans in a room has been shown by Saputra \etal \cite{saputra2012indoor} who juxtaposed two calibrated Kinects at $5m$ distance next to each other. Although the projection cones do not interfere with each other, the authors provide a detection error of human position of $10 cm$.
\section{Interference in Multiple RGB-D sensor Setups}
\label{interference}
Following the work of Berger \etal \cite{berger2011markerless}, where external hardware shutters are used for mitigating interference between concurrently projecting sensors as described in detail by Schroeder \etal \cite{schroder2011multiple}, Maimone and Fuchs \cite{maimone2012reducing} introduce motion platforms that pitch each Kinect with the Kinect that the own structured light pattern remains crisp in the IR stream while the other patterns
appear blurred due to the angular motion of the camera. The depth map is realigned with the recorded egomotion from the inertial sensors included in the Kinect. It is noteworthy that they also managed to deblur the RGB-image using the Lucy-Richardson method. In a more generic approach Butler \etal \cite{butler2012shake} vibrate the camera arbitrarily. 
In a rather invasive approach Faion \etal \cite{faion2012intelligent} manage to toggle the projector subsystem to perform measurements similar to Schroeder \etal \cite{schroder2011multiple}. They use Bayesian state estimator to intelligently schedule which sensor is to be selected for the next time frame. Their maximal reconstruction error denotes $21mm$. Kainz \etal \cite{kainz2012omnikinect} describe an elaborate setup for eight Kinects mounted on vibrating rods and one freely moving Kinect suitable for various applications, such as motion capturing and reconstruction. All vibrating rods were administered by a parallel circuit at slightly different frequencies. They do not give a quantitative analysis of the reconstruction error but provide qualitative figures of the reconstructed mesh.

\section{Conclusion}
In this state-of the art report we have shown that, counter-intuitively, it is possible to use several Kinects in one capturing setup.
Although each device projects the same pattern at the same wavelength into the scene and consequently contributes to confusion in processing the raw IR-data, several approaches, ranging from hardware fixes over intelligent software algorithms for mitigation to placing the Kinects such that the scene content acts as an occluding surface between each projection cone, have been discussed. The applicational context varied between motion capturing, the original purpose of the Kinect sensor, over scene reconstruction to tracking and recognition. With the advent of Kinect One and the change in underlying technology it will be possible to setup multiple sensors in one capturing scenario more conveniently, but the authors predict that in the next years there will still be challenges for multiple RGB-D sensors relying on the emission of light to be addressed by the community. 

\section*{Appendix}
We want to thank the anonymous reviewers for their comments. We thank Yannic Schroeder for providing us with the image material for Figure 1.
{\small
\bibliographystyle{ieee}
\bibliography{egbib}
}
\begin{table*}
\centering
\begin{tabular}{|p{3cm}|p{3cm}|p{3cm}|p{3cm}|p{3cm}|}
\hline
\textbf{Author}&\textbf{Context}& \textbf{Number Of RGB-D sensors in setup} & \textbf{Accuracy} & \textbf{Calibrated}\\
\hline
\hline
Asteriadis \etal \cite{asteriadis2013estimating} &Motion Estimation&3&not specified&Yes\\
\hline
Berger \etal \cite{berger2011markerless} &Motion Estimation&4&Reprojection Error Of $1.7 px$&Yes\\
\hline
Fuhrmann \etal \cite{fuhrmannmulti}&Motion Estimation&3&Deviation of $2-3 cm$&Yes\\
\hline
Hossny \etal~\cite{hossny2012low}&Motion Estimation&2&not specified&Yes (The authors provide a new autocalibration algorithm)\\
\hline
Santhanam \etal \cite{santhanam2011th}&Motion Estimation&4&Deviation of $3mm$&Yes\\
\hline
Wilson \etal \cite{wilson2010combining}&Motion Estimation&3&not specified&Yes\\
\hline
Ye \etal \cite{ye2013free}&Motion Estimation&not specified&not specified&No\\
\hline
Zhang \etal \cite{zhang2012real}&Motion Estimation&2&Deviation of $20 cm$&Yes\\
\hline
\hline
Alexiadis \etal \cite{alexiadis2012reconstruction}&Mesh Reconstruction&4&Reprojection Error Of $0.8 px$&Yes\\
\hline
Berger \etal \cite{berger2011capturing} and
Berger \etal \cite{berger2013a}&Mesh Reconstruction&3&not specified&Yes\\
\hline
Macknojia \etal \cite{macknojia2013calibration}&Mesh Reconstruction&5&Deviation of $2.5 cm$ at $3m$ distance&Yes\\
\hline
Lo \etal \cite{lothree}&Mesh Reconstruction&2&not specified&Yes\\
\hline
Nakara \etal \cite{nakamuramultiple}&Mesh Reconstruction&2&Deviation of $3\%$ at $90^\circ$ spacing&No\\
\hline
Nakazawa \etal \cite{nakazawa2012dynamic}&Mesh Reconstruction&4&not specified&Yes\\
\hline
Naveed \cite{ahmedsystem}&Mesh Reconstruction&6&not specified&Yes\\
\hline
Olesen \etal \cite{olesen2012real} &Mesh Reconstruction&3&$60\%$inlier at $8px$ Texlet spacing&Yes\\
\hline
Pancham \etal \cite{pancham2012mapping}&Mesh Reconstruction&$2+$&not specified&No\\
\hline
Rafibakhsh \etal \cite{rafibakhsh2012analysis}&Mesh Reconstruction&2&Deviation of $3.49cm$&Yes\\
\hline
Sumar \etal \cite{sumarfeasability}&Mesh Reconstruction&2&Reprojection Error Of $5 px$&No\\
\hline
Wang \etal \cite{wang20123d}&Mesh Reconstruction&2&not specified&Yes\\
\hline
\hline
Caon \etal \cite{caon2011context}&Recognition&3&not specified&Yes\\
\hline
Satta \etal \cite{sattareal}&Recognition&2&not specified&No\\
\hline
Satyavolu \etal \cite{satyavolu2012analysis}&Recognition&5&Deviation of $3 cm$&Yes\\
\hline
Saputra \etal \cite{saputra2012indoor}&Recognition&2&Deviation of $10 cm$&Yes\\
\hline
Susanto \etal \cite{susanto20123d}&Recognition&5&Deviation of $13 cm$&Yes\\
\hline
\hline
Butler \etal \cite{butler2012shake}&Interference&2 and 3&Deviation of up to $3 cm$&Yes\\
\hline
Faion \etal \cite{faion2012intelligent}&Interference&4&Deviation of $21 mm$&Yes\\
\hline
Kainz \etal \cite{kainz2012omnikinect}&Interference&8&not specified&Yes\\
\hline
Maimone and Fuchs \cite{maimone2012reducing}&Interference&6&Deviation of $2 mm$&No\\
\hline
Schroeder \etal \cite{schroder2011multiple} and 
Berger \etal \cite{berger2011markerless} &Interference&4&Reprojection Error Of $1.7 px$&Yes\\
\hline

\end{tabular}
\vspace{.5cm}
\caption{\label{overviewtable} An Overview over different publications including multiple RGB-D sensors. The table lists for each publication the amount of employed sensors, the context pof application, the accuracy and whether the sensors where calibrated to a common world space. Note, that the specification of accuracy varies with the context of application between the mean deviation of a reconstructed 3d position from the original position in meters and the reprojection error in pixels or percentage into the camera.}
\end{table*}
\end{document}